# PISHYAR: A Socially Intelligent Smart Cane for Indoor Social Navigation and Multimodal Human-Robot Interaction for Visually Impaired People

Mahdi Haghighat Joo[1], Maryam Karimi Jafari[1], Alireza Taheri[1*]

[1]Social and Cognitive Robotics Laboratory, Sharif University of Technology, Tehran, Iran

* Corresponding author: artaheri@sharif.edu , Tel: +982166165531

**Abstract**

Visual impairments pose significant challenges to safe mobility and situational awareness for visually impaired and low-vision individuals, particularly in socially dynamic indoor environments. This paper presents PISHYAR, a socially intelligent smart cane designed by our group to combine socially aware navigation with multimodal human-AI interaction to support both physical mobility and interactive assistance. The system consists of two components: (1) a social navigation framework implemented on a Raspberry Pi 5 that integrates real-time RGB-D perception using an OAK-D Lite camera, YOLOv8-based object detection, COMPOSER-based collective activity recognition, D* Lite dynamic path planning, and haptic feedback via vibration motors for tasks such as locating a vacant seat; and (2) an agentic multimodal LLM-VLM interaction framework that integrates speech recognition, vision language models, large language models, and text-to-speech, with dynamic routing between voice-only and vision-only modes to enable natural voice-based communication, scene description, and object localization from visual input. The system is evaluated through a combination of simulation-based tests, real-world field experiments, and user-centered studies. Results from simulated and real indoor environments demonstrate reliable obstacle avoidance and socially compliant navigation, achieving an overall system accuracy of approximately 80% under different social conditions. Group activity recognition further shows robust performance across diverse crowd scenarios. In addition, a preliminary exploratory user study with eight visually impaired and low-vision participants evaluates the agentic interaction framework through structured tasks and a UTAUT-based questionnaire reveals high acceptance and positive perceptions of usability, trust, and perceived sociability during our experiments. The results highlight the potential of PISHYAR as a multimodal assistive mobility aid that extends beyond navigation to provide socially interactive support for such users.

**Keywords:** smart cane, socially assistive technology, visually impaired and low-vision users, social navigation, group activity recognition, large language models, human-robot interaction

# 1 Introduction

In recent years, robotic systems have been rapidly developing to assist and enhance various aspects of human life, including delivery and service roles [1], industries [2] and healthcare [3]. One important application of these technologies is improving daily life for individuals with special needs. Social robotics has shown strong potential in this area, including children with autism [4, 5], sign-language recognition [6], and lip-reading training for hearing-impaired individuals [7]. Among these groups, visual impairments present significant challenges and make routine tasks difficult for affected individuals.

Visual impairments remain a growing global concern, with over 2.2 billion individuals affected worldwide according to the World Health Organization. Among them, at least one billion people suffer from moderate to severe vision loss due to preventable or unaddressed conditions such as cataracts, uncorrected refractive errors, diabetic retinopathy, glaucoma, trachoma, and presbyopia [8]. These impairments significantly reduce quality of life and limit personal independence. As the global population ages and the prevalence of chronic diseases rises, the need for advanced assistive technologies that support independent mobility and environmental awareness has become increasingly urgent.

Traditional mobility aids, such as white canes and guide dogs, while accessible and widely used, are often insufficient in complex and dynamic environments [9]. These tools typically lack the capability to provide detailed environmental context or real-time feedback, which are essential for safe and autonomous navigation. In response, there has been a growing focus on the development of intelligent assistive systems that integrate modern technologies such as sensors, embedded cameras, and real-time processing algorithms, to deliver actionable information to users through auditory, vibratory, or multimodal feedback mechanisms [10]. Beyond physical navigation support, assistive systems for visually impaired users increasingly emphasize interaction and communication. Human-robot interaction (HRI) plays a critical role in this context, as it enables not only functional assistance but also socially meaningful interactions between users and assistive devices [11].

Recent advancements in Large Language Models (LLMs) and Vision Language Models (VLMs) have opened new possibilities for developing interactive and socially aware assistive technologies for visually impaired users [12]. In contrast to traditional rule-based systems, LLMs enable natural language communication, and VLMs allow the system to connect visual information with user requests. These features enable assistive devices such as smart canes to go beyond simple obstacle detection and provide richer services, including scene understanding, object localization, and conversational guidance.

In this paper, we present **PISHYAR**, our developed socially intelligent smart cane designed to support visually impaired users through both social navigation and interactive assistance. The system is named **PISHYAR** (Persian: پیشیار), reflecting its dual role as a "pioneer" that guides the user and as a predictive assistant that anticipates upcoming environmental elements in front of the user. The proposed system provides two complementary but independently developed system-level contributions: 1) a socially aware navigation framework that integrates real-time perception, object detection, group activity recognition, and path planning for locating and navigating to vacant seats in indoor environments, and 2) an agentic multimodal interaction framework based on LLM-VLM cooperation that enables natural voice-based communication, scene description, and object localization. The system is evaluated through a combination of simulation-based tests and real-world experiments to assess navigation accuracy and socially aware behavior. In addition, a preliminary exploratory user study with eight visually impaired participants is conducted to initially investigate the multimodal interaction framework using a UTAUT-based questionnaire, which focuses on usability, perceived usefulness, perceived ease of use, user acceptance, intention to use, trust, and perceived social intelligence. An overview of the proposed system is shown in Fig. 1.

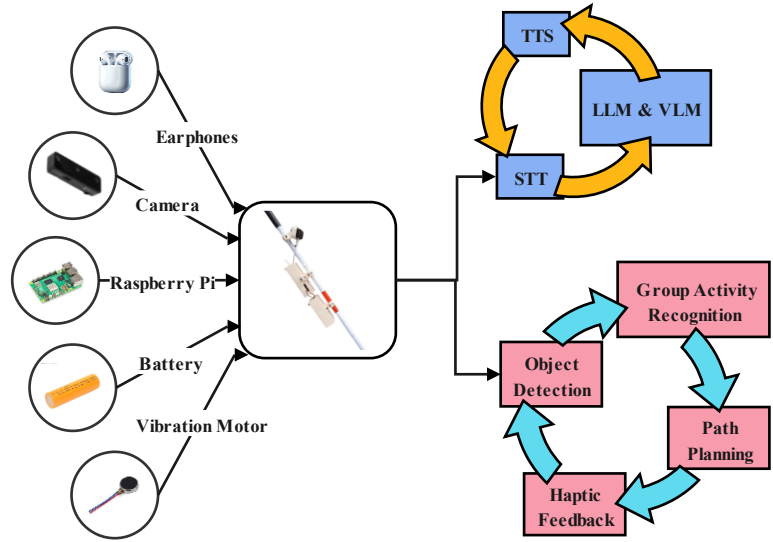

**Fig. 1** An overview of the study: PISHYAR hardware design and its capabilities (for visually impaired individuals).

## 2 Related Works

Over the past two decades, significant efforts have been made to develop smart assistive technologies for visually impaired individuals. Among these, smart canes have attracted the most attention and can generally be categorized based on their core function, such as obstacle detection, navigation assistance, or support for daily activities. These categories are further subdivided based on sensor type and intended environment (indoor vs. outdoor) [13].

Several studies have focused on obstacle detection. For example, Masud et al. [14] proposed a low-cost, wearable smart cane using Raspberry Pi 4B, camera, ultrasonic sensor, and Arduino, employing Viola-Jones and TensorFlow Object Detection for accurate scene classification and obstacle detection. Alashjaee et al. [15] introduced the HAODVIP-ADL system, an attention-driven deep learning approach combining YOLOv10, CapsNet, InceptionV3, and MHA-BiLSTM, achieving highly accurate (99.74%) object detection and safe navigation support while enabling family monitoring via a web-based application. Bauer et al. [16] developed a low-cost, non-intrusive wearable system that estimates depth maps from color images, detects urban objects, and conveys 3D and semantic obstacle information via haptic or spoken feedback, achieving 87.99% mean accuracy in real-world outdoor navigation. Muktha et al. [17] developed a smart cane equipped with ultrasonic sensors to detect nearby obstacles and provide haptic feedback via a bidirectional vibration motor before contact occurs. This cane was also able to assist visually impaired users with navigation using a voice assistant.

Some studies have focused on supporting daily activities. Abdusalomov et al. [18] proposed an AI-based vision system using YOLOv5m for early indoor fire detection, providing real-time alerts to blind and visually impaired individuals with high speed and accuracy. Dang et al. [19] developed a deep learning-based system for kitchen environments, which offers real-time object detection and interactive guidance via speech to enhance accuracy, autonomy, and safety in complex settings.

Most recent developments target navigation and obstacle avoidance. Slade et al. [20] developed the Augmented Cane, a low-cost, sensor-equipped white cane that provides steering assistance, obstacle avoidance, and object recognition, improving walking speed by up to 35% and enhancing mobility, confidence, and safety. Feltner et al. [21] combined a depth camera, Raspberry Pi, Kinect sensor, and vibration motors to provide tactile feedback for obstacle awareness. Agrawal et al. [22] created a perceptual robotic cane with haptic guidance, enabling users to identify and approach preferred seats in public spaces. The system used RGB-D cameras, inertial measurement units (IMUs), vibration motors, and portable processors, integrating SLAM and goal-oriented planning for accurate seat recognition. Mai et al. [23] introduced a LiDAR and vision-based smart cane capable of real-time obstacle detection and localization. The system employed 2D LiDAR, RGB-D cameras, GPS, and a Jetson Nano processor. Object recognition was achieved using YOLOv5, and SLAM was used for environment mapping. Real-world evaluations showed pedestrian detection accuracy of 84.6%, zebra-crossing detection at 71.8%, and overall object recognition accuracy of 61.2% across 86 object types.

Finding vacant seats is an important need for visually impaired individuals [24]. This has motivated research toward goal-aware assistive systems that navigate users toward specific targets. However, prior works do not consider social etiquette during navigation—a critical factor for user comfort and privacy.

With recent advances in LLMs, their use in assistive technologies for visually impaired individuals has gained increasing attention. Recent studies have explored LLM- and VLM-based approaches for tasks such as object detection, scene description, and navigation assistance [25, 26]. Video-based LLMs such as GPT-4o [27] have demonstrated strong performance in real-time tasks related to home and social environments, which indicates their potential for interactive assistance. Other works have shown that these models can generate navigation instructions and contextual guidance [25], while safety-aware systems using GPT-4V [26] highlight the ability of VLMs to assess environmental risks in real time.

Several recent works have attempted to integrate LLMs and VLMs into practical assistive systems for visually impaired users. MASA [28] proposed a wearable system using GPT-4o to interpret camera input and user queries through an agentic pipeline, with results showing high usability in user studies. NAVISIGHT [29] combined YOLOv11n and LLaVA for object recognition and scene description, although it did not include user evaluations. Nevasekar et al. [30] introduced a robot-assisted navigation system using YOLOv8 and LLaVA with retrieval-augmented generation (RAG) for visual question answering. PathFinder [31] employed LLMs and VLMs together with monocular depth estimation for navigation, and measured user engagement and response time. Hao et al. [32] proposed a system that uses a VLM for scene understanding and hazard detection, with responses generated by a Vicuna-based LLM, but without interactive navigation experiments. Although these studies show the potential of LLMs and VLMs for assistive tasks, they often focus on specific tasks and do not fully address socially aware navigation or integrated interaction in smart canes. Unlike previous methods, our designed robot, PISHYAR, integrates socially aware navigation and agentic multimodal interaction within a single smart cane platform, evaluated through both technical experiments and user studies.

**3 Hardware Design**

This section presents the design and implementation of PISHYAR hardware, which constitutes the first core component of the proposed system. The hardware is designed to support environmental perception, onboard computation, and intuitive user feedback, which enables both social navigation and interactive functionalities. PISHYAR integrates a depth-sensing camera (OAK-D Lite), a Raspberry Pi 5 as the central processing unit, vibration motors for haptic feedback, a power supply module, and a custom mechanical cane structure.

One of the key components of the system is the depth-sensing camera, whose primary function is to acquire information from the environment and calculate the distances of objects relative to the user. The OAK-D Lite camera was selected as the primary sensor due to its balance of depth measurement accuracy, real-time processing capability, ease of integration, and low power consumption. In addition to providing high-resolution RGB-D data, this camera is equipped with an internal neural processing unit (Myriad X VPU), which allows computer vision algorithms to be executed independently without reliance on an external processor. This feature reduces the computational load on the central processing unit and improves system response time. Moreover, the camera includes a built-in six-degree-of-freedom IMU, eliminating the need for a separate motion sensor module and contributing to a more lightweight design.

For the central processing unit, criteria such as high computational power, support for multiple inputs and outputs, simultaneous multi-sensor data processing, and compatibility with machine learning algorithms were considered. Accordingly, the Raspberry Pi 5 was chosen as the processing core. With its powerful processor and adaptable communication interfaces, this board can simultaneously manage processes such as navigation, sensor control, and real-time feedback delivery.

The power supply module uses two 18650 lithium-ion batteries connected in series to provide 7.4 V, managed by a two-cell battery management system (BMS) for balanced charging and protection. A power switch enables manual control of the system. The output is regulated to 5 V using an XL4015 DC–DC buck converter and delivered to the Raspberry Pi 5 via a USB Type-C port.

To provide user feedback, three vibration motors were installed on the cane handle, each corresponding to a specific direction (left, right, and forward). These motors are controlled by the Raspberry Pi 5 and are activated for one second depending on the required direction of movement. This haptic feedback system provides a simple way for visually impaired users to receive spatial information.

Mechanical design and component placement were fully performed using SolidWorks 2023. Special attention was given to handle ergonomics, weight distribution, and ease of maintenance. After completing the design, components were manufactured using FDM 3D printing, which not only reduced production costs but also allowed rapid design modifications and customization. The initial version of the cane, consisting of only the basic structural elements, weighed approximately 500 grams. After integrating the camera, processing unit, vibration motors, battery, and other electronics, the total weight of the smart cane reached around 1 kg. The total cost of the entire assembly is approximately 400 USD. Fig. 2 illustrates an overall view of PISHYAR.

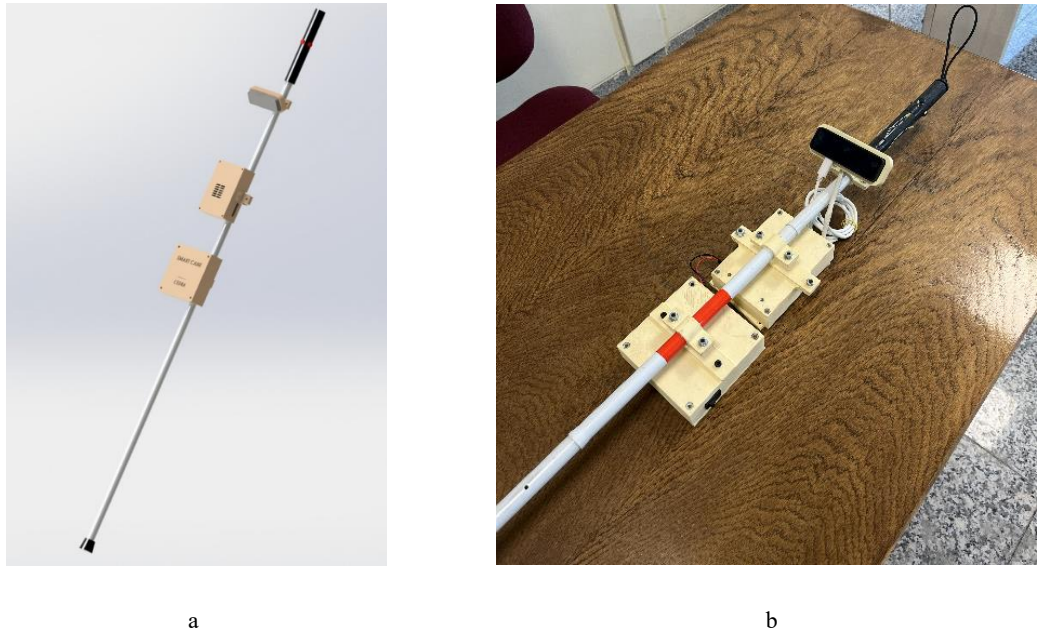

a                                                                                  b

**Fig. 2** The developed PISHYAR smart cane prototype: (a) SolidWorks 3D model of the cane structure, and (b) the implemented physical prototype.

## 4 Social Navigation for Indoor Vacant Seat Finding

This section presents the second core component of the proposed system, focusing on the design and implementation of socially aware indoor navigation for vacant seat finding. The goal of this component is to enable safe and socially appropriate indoor navigation by combining object detection, path planning, and group activity recognition. This part of the system primarily relies on the depth-sensing camera, the onboard processing unit, and the haptic feedback actuators. This section first introduces the system modules for social navigation, then describes the execution process and evaluation methods.

### 4.1 Social Navigation System Architecture

#### 4.1.1 Object and Obstacle Detection for Social Navigation

As part of the proposed system, this section focuses on socially aware navigation in indoor environments, with the specific goal of identifying vacant seats and guiding the user toward them. To achieve this goal, a high-accuracy, real-time obstacle detection model was required. After evaluating several options, the YOLOv8n architecture was selected as the primary model which can be executed on the camera processing unit [33]. Trained on the COCO dataset, this model is capable of recognizing objects from 80 different classes across a large-scale annotated dataset.

After detecting objects in the environment, target selection was performed based on two criteria: the unoccupied seat and the shortest distance to the user. Distances were calculated using the Euclidean metric. Once the target was determined, the coordinates of the seat and obstacles were provided to the path planning algorithm.

**4.1.2 Path Planning Algorithm**

Another key component of the proposed system is the path planning algorithm. The purpose of the algorithm is to provide a safe and shortest path while respecting all constraints of the problem. Additionally, given the coordinates of the start point, destination, and obstacles, the algorithm must be able to generate an optimal path in real time, and instantly recalculate an alternative route if the previous path becomes blocked. Accordingly, the D* Lite algorithm, a grid-based optimized method suitable for dynamic environments [34], was employed. To simplify navigation, the user's permissible directions were limited to three main options: forward, left, and right. After selecting the target and sending the coordinates, the path proposed by the algorithm was transmitted to PISHYAR, and user guidance was executed using the three primary directions.

**4.1.3 Group Activity Recognition Using COMPOSER**

To train and evaluate the group activity recognition model, the standard Collective Activity Dataset was used, which was first collected and annotated in 2009 by Choi and colleagues [35]. This dataset includes five social activity classes: queuing, talking, waiting, walking, and crossing. In our study, Given the high similarity between "crossing" and "walking," these two activities were considered as a single class, referred to as "walking". In total, it contains 2,543 ten-frame clips with a resolution of 480×640 pixels. Videos were recorded from different angles to ensure spatial diversity and introduce realistic visual challenges. Each clip is accompanied by detailed labels, including the type of activity, individuals' orientations, and their positions within each frame.

Recognizing collective activity is one of the most challenging tasks in video understanding due to the need to capture spatiotemporal relationships among people and objects. Traditional approaches based on RGB images or optical flow, besides being heavily dependent on background and scene conditions, often face issues such as environmental bias and privacy concerns. To address these challenges, this work adopts the COMPOSER architecture proposed by Zhou et al. [36], a multiscale transformer-based model that performs group activity recognition using only 2D/3D human key-point coordinates rather than raw RGB images. By modeling interactions between individuals at multiple semantic levels, COMPOSER enables reliable recognition of collective activities while reducing sensitivity to background variations and preserving user privacy. In this study, COMPOSER is used as a pretrained framework and retrained on the Collective Activity Dataset for the target activity classes.

The COMPOSER framework represents collective activities using a hierarchical, multi-scale tokenization scheme that encodes human motion and social context from body key points to group-level interactions, while also incorporating object information. A global clip-level token aggregates contextual information across all scales to enable robust recognition of group activities. To extract the 17 key body points required as input to the COMPOSER model for activity recognition, this research utilized the MediaPipe library [37]. Although this approach is different from the method originally used in the COMPOSER architecture, it was chosen because it is simpler and has a faster processing speed, which makes it more suitable for real-time deployment.

For model configuration, the hidden size for token representation was set to 128, shared across query, key, and value vectors in all transformer encoders; while the hidden dimension of the MLP layers was fixed at 1024. The architecture employs two multiscale transformer blocks, each containing separate encoders for the different scales. The number of attention heads was set to 2, 8, 2, and 2 for the four scales, respectively; with corresponding dropout rates of 0.5, 0.2, 0.2, and 0. ReLU was chosen as the activation function in the MLP layers. To improve generalization, several data augmentation techniques were applied, including horizontal flipping, horizontal and vertical shifting, and random actor dropout; additionally, small random noise (±1 pixel) was added to the coordinates during certain steps. For optimization, the Adam optimizer was used with an initial learning rate of 0.0005, reduced to 0.0001 after 40 epochs, and a batch size of 384.

**4.2 Social Navigation Execution Process**

The path planning process in this system is carried out through a set of structured and interconnected steps to ensure the visually impaired user is guided safely while adhering to social navigation etiquette in public spaces. These steps are shown in Fig. 3. The execution steps, incorporating group activity recognition, are defined as follows:

**1. Initial User Positioning:** The user holds the cane and is given 10 seconds to position themselves correctly and adjust the cane at a standard angle (approximately 45° relative to the horizontal plane). This time allows stabilization and prevents initial data collection errors.

**2. Data Acquisition and Initial Analysis:** After the initial period, the RGB-D camera begins capturing data. The first 10 frames are sent to both the obstacle and object detection model (utilizing YOLO) and the group activity recognition architecture. To improve accuracy, distance data from these frames are compared. If an object's distance deviates more than 10% from the mean of the 10 frames, it is identified as an outlier and removed. If a seat is initially selected as the target and is not occupied by another person, this choice remains fixed until the end of the path. Even if better seats are detected later, the originally selected target does not change.

**3. Group Activity Recognition:** Those initial 10 frames are sent to be processed by the group activity recognition model. If no collective activity is detected, the system proceeds to standard path planning. In this case, the system skips social condition adaptation and performs path planning using only obstacle detection and goal information. The detection of collective activities is performed only at the beginning of the movement, under the assumption that no new human groups enter the immediate navigation area during execution.

**4. Social Condition Adaptation:** The environment is divided into grid cells for path planning, and detected human groups are expanded into surrounding cells, which are then treated as obstacles. If a collective activity (e.g., walking, talking, queuing) is identified, path planning decisions are adjusted according to the activity type:

- **Walking Activity:** The system waits 10 seconds for the path to clear, indicated to the user via three short activations of the forward-direction vibration motor. If the group persists, normal path planning resumes.
- **Talking Activity:** The area occupied by the talking group is marked as non-navigable, and two adjacent cells are blocked for safe distancing.
- **Queuing or Waiting Activity:** The entire group is treated as a restricted area, with one extra row adjacent to the group also marked as non-navigable to maintain minimum social distance.

**5. Object Analysis and Path Planning Input:** Following initial data filtering and collective activity recognition, detected valid objects and obstacles, along with their distances from the cane, are sent to the D* Lite algorithm to compute the optimal path.

**6. User Guidance Start:** The initial direction (left, right, or forward) is communicated to the user through the vibration motors based on the calculated path, considering any social condition adjustments.

**7. Movement Monitoring and Position Verification:** After the user moves, their displacement is analyzed by comparing data from 10 consecutive frames. The measurement metric uses 10 randomly selected pixels in the image to estimate displacement consistency across frames. Step length is set to 40 cm for blind users, slightly shorter than the average sighted step, to improve accuracy and safety.

**8. Next Direction Announcement:** If the movement and direction are verified with acceptable accuracy (less than 10% deviation from predictions), the next direction is calculated and communicated to the user.

**9. Target Review and Alternative Path Planning:** At each step, the algorithm checks if the target seat has been occupied. If so, a new alternative path is suggested.

**10. Cumulative Error Correction:** To mitigate errors from sensor inaccuracies or model drift, the user's calculated position is compared with the reference map after every three steps, and corrected if significant deviation is detected.

**11. Iteration Until Reaching the Goal:** This process loops until the user reaches the final target.

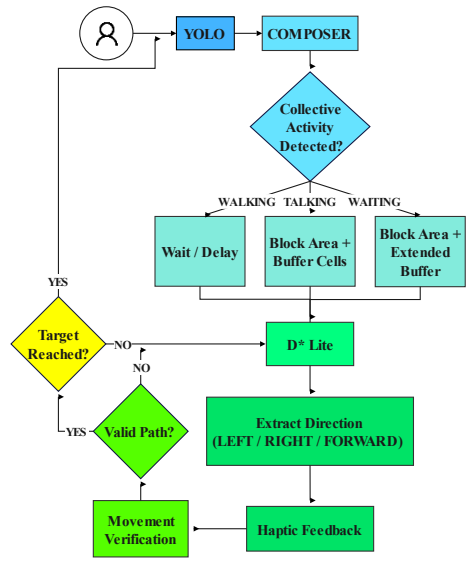

**Fig. 3** Social navigation workflow.

### 4.3 Social Navigation Evaluation Methodology

### 4.3.1 Simulation-Based Evaluation

In the first step, the combination of the obstacle detection model and the path planning algorithm was evaluated using simulation. For this simultaneous evaluation, Webots robotic simulation platform [38] was used. In this environment, a humanoid model equipped with the smart cane was implemented. The cane was fitted with an RGB-D camera and a six-degree-of-freedom IMU. The camera fed image frames to the YOLO model to identify objects and estimate their distances relative to the user, while the IMU recorded the user's movement direction in real time. To increase the safety of blind users' movements, the simulated environment was divided into cells measuring 40 cm each, slightly shorter than the average step length of sighted individuals. This evaluation was conducted solely in the presence of static obstacles and human figures; where the number, type, and spatial arrangement of obstacles and objects were randomly varied in each test to generate diverse environmental layouts, allowing the system's behavior to be examined under controlled conditions. The designed model was evaluated across ten such scenarios to verify the accuracy and reliability of the obstacle detection and path planning modules, and to confirm their correct integration and coordinated operation prior to real-world deployment. The primary success criterion in this simulation stage was the system's ability to guide the user toward the target seat without collisions.

### 4.3.2 Real-World Evaluation of Group Activity Recognition

In the next step, the trained COMPOSER model was evaluated on real-world data obtained from the camera mounted on PISHYAR, besides its evaluation on the test portion of the training dataset. For this purpose, after the required preprocessing, ten consecutive frames from the camera's output were provided as input to the model to perform collective activity recognition.

For performing this evaluation, experiments were designed and conducted across 20 different scenarios. In these experiments, each group of collective activities was present in the environment at least five times to examine the influence of position and viewing angle variations on model accuracy. In order to replicate real-world difficulties, the test conditions were created to be as varied as possible in terms of camera angles and environmental settings. Fig. 4 presents an example of images captured during these experiments, illustrating the variety of environments and test conditions.

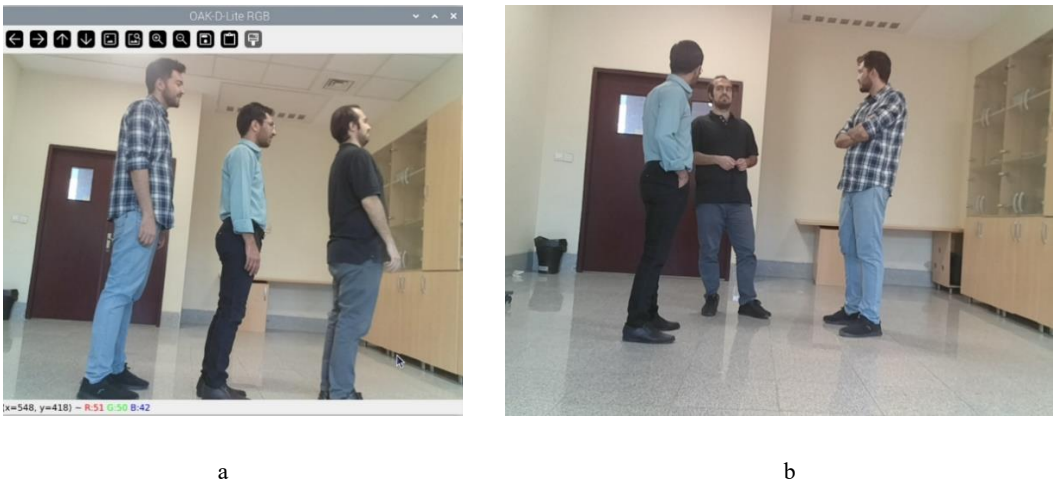

a　　　　　　　　　　　　　　　　　　　　b

**Fig. 4** Example images captured by PISHYAR for group activity recognition: (a) queuing activity, and (b) talking activity.

### 4.3.3 Field Evaluation of the System Across Scenarios

In the last step of evaluation of the whole system, including the hardware and the algorithms, we decided to perform field evaluation across some scenarios. The evaluation process was conducted in an indoor environment measuring approximately 3 meters in width and 5 meters in length, which was suitable for flexible arrangement of obstacles and positioning of human groups at various angles and locations. The purpose of this setup was to accurately simulate real-world conditions so that the performance of PISHYAR could be thoroughly tested under a wide range of environmental and behavioral variations. Notably, the evaluations were performed by a sighted individual with their eyes closed to replicate the experience of a visually impaired person. This approach was adopted for safety reasons during early-stage testing and to ensure controlled and repeatable evaluation conditions. To assess overall system performance, 15 experimental scenarios were designed and executed, divided into three categories:

- **Scenario 1, Static Obstacles without Human Groups:** Only fixed obstacles such as tables, chairs, and other objects were present, and the goal was to reach an empty seat without collisions. This scenario served as the initial evaluation of system stability. Each scenario was repeated three times with random variations in obstacle positions and target seat location to assess adaptability.

- **Scenario 2, Static Obstacles with Target Changes:** Similar to Scenario 1, but during movement, the target seat was suddenly occupied by another person. The system had to quickly identify an alternative path and guide the user to another empty seat. This scenario was also repeated three times with randomized obstacle and target placements to evaluate system flexibility and replanning capability.

- **Scenario 3, Social Path Planning with Human Groups:** The most complex and research-focused scenario involved the presence of human groups in the environment. The system needed to detect the group, analyze their activity type, and navigate in a way that maintained social etiquette, such as keeping proper distance and avoiding disturbance. This scenario was executed in nine different configurations: three for walking, three for talking, and three for waiting activities. Due to the similarity between waiting and queuing cases, only the waiting class was used for execution.

These scenarios were designed to evaluate all system components, from object detection and path planning to social activity recognition and feedback delivery, under conditions that closely simulate real-world operation. To comprehensively assess the system's performance, the evaluation was based on three key and essential criteria: 1) successfully reaching the designated target, 2) avoiding collisions with any obstacles along the path, and 3) strictly observing social etiquette and interactions in the presence of human groups. In each scenario, the system's performance was considered acceptable only if all three criteria were met simultaneously. Fig. 5 shows an example from one of the evaluated scenarios, illustrating the diversity of positions and testing conditions.

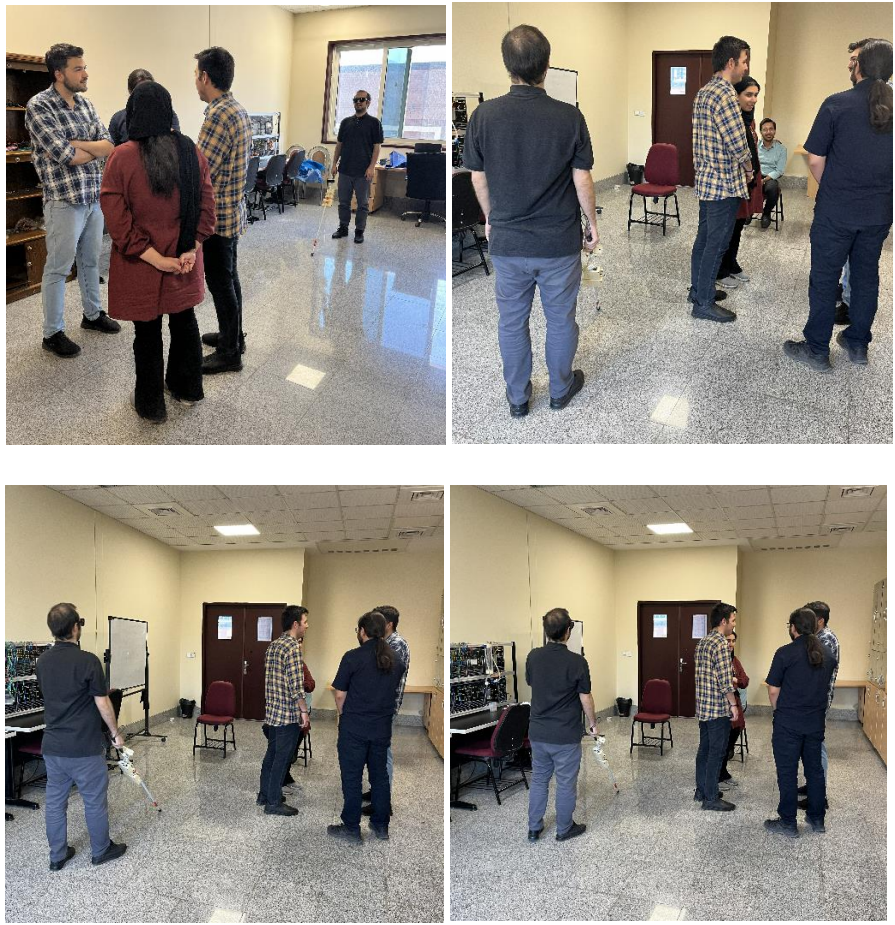

**Fig. 5** Example of a social navigation evaluation scenario.

## 5 LLM- and VLM-Based Social Interaction Framework

This section presents the third core component of the proposed system, focusing on the development and evaluation of PISHYAR's social and conversational intelligence. This component integrates LLM and VLM to enable natural dialogue and visual scene understanding. Its implementation relies on key hardware elements of the smart cane, including the Raspberry Pi, depth-sensing camera, power supply, physical cane structure, and earphones for audio feedback. This section first introduces the multimodal interaction architecture and system modules, then describes the agentic interaction framework, and finally presents the user study design and evaluation metrics.

### 5.1 Multimodal Social Interaction Architecture

The proposed multimodal interactive system consists of several interconnected modules that collaboratively enable natural and multimodal communication between the user and PISHYAR. These modules are described as follows:

#### 5.1.1 Speech Recognition Module (STT and KWS)

A speech-to-text (STT) module is used to convert the user's spoken input into text. Google Speech Recognition was employed due to its support for the Persian language, which is accessible through the Python *SpeechRecognition* library. In addition, a keyword spotting (KWS) module was integrated to activate the STT system only when the user explicitly calls the system using predefined phrases such as "Hello, PISHYAR" or "PISHYAR, my friend." To achieve this, the Vosk small Persian STT model was used to perform KWS. Vosk is an open-source toolkit that works well on local devices without an online connection and allows accurate voice recognition in a variety of languages, including Persian. Because the Vosk small model is lightweight and effective but inaccurate for transcribing long or complex spoken words, it was primarily utilized for keyword spotting.

### 5.1.2 Scene Description Module (VLM)

The system utilizes GPT-4o as a VLM for indoor scene description. Whenever the user requests a description of the surrounding environment, an image captured by the OAK-D camera is sent to the VLM, and the generated textual description is returned to the system.

### 5.1.3 Object Detection Module (YOLO)

Object detection is performed using the YOLOv8n model in conjunction with the VLM. This module detects objects in the environment and estimates their distance from the user using depth information from the RGB-D camera, which is converted into approximate step counts. This module runs the YOLO on the camera's internal neural processing unit instead of the Raspberry pi.

### 5.1.4 Natural Language Dialogue Module (LLM)

GPT-4o is employed as the core LLM responsible for managing user interaction. The LLM processes the user's textual input and determines whether visual perception is required. It has access to the outputs of both the VLM and the YOLO module to generate context-aware responses. Additionally, it is capable of answering general-purpose questions independent of visual input.

### 5.1.5 Date and Time Provider Module

This module uses Python's built-in tools to get the current system date and time. It then converts them into the Persian (Jalali) calendar when requested by the user.

### 5.2 Agentic Interaction Framework

The interaction framework is designed as an agentic system, in which the LLM acts as a decision-making agent that dynamically routes user requests and invokes perception and dialogue modules based on the inferred intent. The reason for using an agentic interaction framework is to enable context-aware control over multiple system modules and reduce unnecessary computation. By treating the LLM as an agent, the system can interpret user intent and dynamically decide which perception and dialogue components should be activated, rather than executing all modules continuously. This design is particularly suitable for assistive systems, where efficient resource usage is critical. The agentic workflow of PISHYAR is shown in Fig. 6. All prompts used for the LLM and VLM components are provided in the Appendix.

According to Fig. 6, interaction with the system begins when the user calls "PISHYAR," which activates the STT module for a command input. The user's speech is converted into text and forwarded to an LLM-based routing module, referred to as the **Interaction Mode Router (IMR)**. Based on the content of the user's request, this router selects one of two operational modes: ***VISION*** mode, which involves camera-based perception for scene description or object detection, and *VOICE* mode, which does not require visual input and is used for general dialogue, including date and time queries. The routing between VOICE and VISION modes enables the camera and vision modules to be activated only when visual input is required. This reduces unnecessary computational load and power consumption on the embedded hardware, which is particularly important for assistive devices. This design introduces a trade-off between responsiveness and computational efficiency, as mode switching may cause slight delays; however, it enables more efficient resource utilization.

In *VISION* mode, a secondary LLM-based router, referred to as the **Vision Task Router (VTR)**, determines whether the user's request corresponds to *SCENE* mode (scene description) or *OBJECT* mode (object detection). The VLM module is utilized in both cases; however, the YOLO module is activated only in OBJECT mode in order to estimate object distances. Upon entering this mode, the camera is initialized and an image is captured. The VLM processes the image to generate a scene description, while the YOLO module operates for a duration of two seconds to detect objects and estimate their distances. This time window was selected to ensure robust detection of all visible objects. Detected objects are filtered to retain unique instances, and estimated distances are converted into step counts. The outputs of the VLM and YOLO modules are then transmitted to the LLM, which synthesizes a final response to the user.

In *VOICE* mode, if the user requests temporal information, the output of the Python date and time provider module is passed to the LLM to generate the response. For all other queries, the LLM generates responses independently. Finally, the LLM's textual output is sent to the TTS module, and the synthesized audio is transferred to the user through earphones.

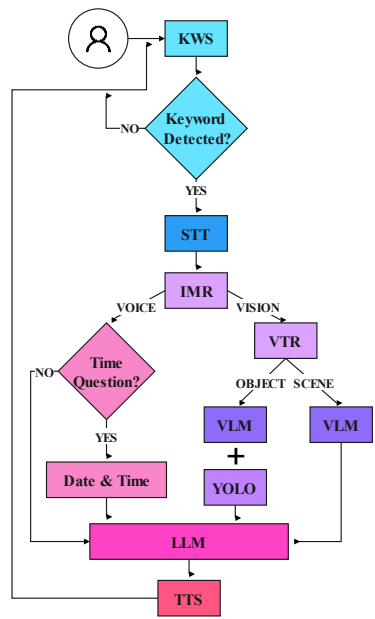

**Fig. 6** The agentic workflow of PISHYAR.

### 5.3 User Study Design and Evaluation Metrics

#### 5.3.1 Experimental Setup

The user study experimental setup consisted of an indoor room environment. Participants initially sat on a chair positioned at the starting point, where instructions were provided and the test began. At a distance of approximately two meters from the starting position, a table and two additional chairs were placed at the opposite end of the room. Several everyday objects were arranged on the table, including a water bottle, a cup, an apple, a spoon, a fork, a book, and a notebook. You can see the setup environment on Fig. 7.

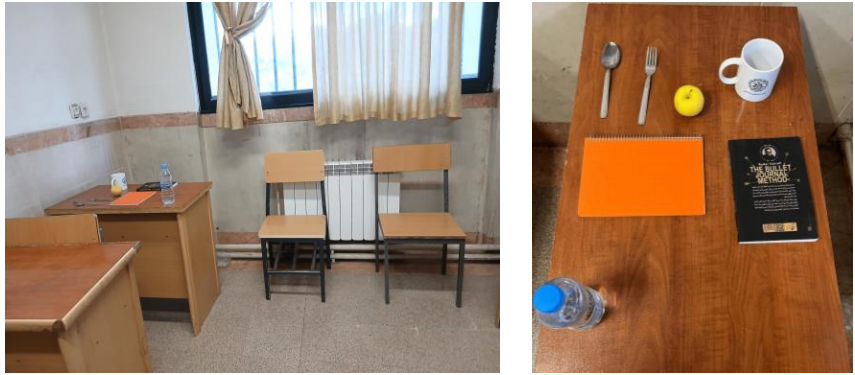

**Fig. 7** Experimental setup: (a) room environment, (b) object arrangement on the table.

#### 5.3.2 Experimental Scenario

At the beginning of each trial, while participants were seated and received verbal instructions about the tasks, they were encouraged to physically explore the smart cane to become familiar with the device. After this introduction, the test session began. The participants were supposed to do at least 4 tasks, and if they wanted, they could ask PISHYAR for any other desired commands in the end.

PISHYAR first introduced itself, and the participant heard the introduction through the earphones. The participant then called PISHYAR, triggering the system. Upon receiving the response "Yes, I am listening" from PISHYAR and through the earphones, the STT module was activated, and the participant asked about the current date and time. The IMR selected the VOICE mode, and the system retrieved the requested information and responded via the TTS module.

In the second task, the participant again called PISHYAR and requested a description of the surrounding environment. The LLM routers selected VISION mode and then SCENE mode. An image captured by the camera was processed by the VLM, and the LLM generated a response based on the scene description. The resulting audio output was delivered through the earphones.

In the third task, the participant asked the system to locate a chair. The system entered VISION and then OBJECT mode. The VLM and YOLO modules generated perception outputs, which were used by the LLM to produce the response. If YOLO successfully detected the chair, the response included an estimated distance in terms of steps;

otherwise, the response was generated based solely on the VLM output. After locating the chair, the participant stood up, walked toward it, and sat down.

In the fourth task, the participant held PISHYAR in one hand, oriented toward the right, and asked the system to locate the water bottle that is on the table to their right. The system again operated in VISION mode, activating both the VLM and YOLO modules, and detected the water bottle on the edge of the table, enabling the participant to reach out and grasp it. After completing this task, the participant was allowed to ask any additional questions of their choice, which were handled by the LLM in VOICE mode. They could also request any other perception task from PISHYAR. Finally, the participant said goodbye to the system, and the session concluded.

All interactions were conducted in the Persian language. The experiment including explanations, doing the tasks, and answering the questions, lasts about 20 to 25 minutes for each participant. The log of the interactions was also captured and saved, which consists what has been transferred between the participant and PISHYAR. A participant interacting with PISHYAR is shown in Fig. 8.

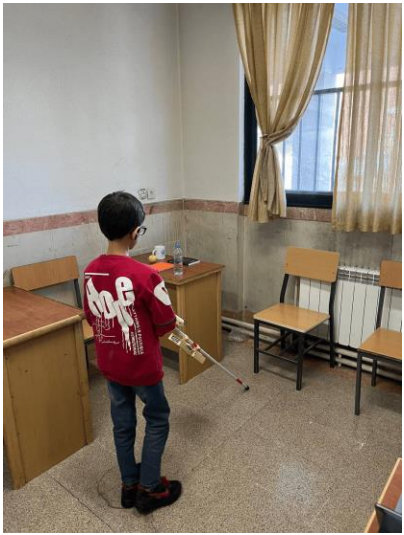

**Fig. 8** A participant interacting with PISHYAR during our user study.

### 5.3.3 Participants
Eight participants were recruited from a local school for visually impaired individuals. The average age was 11.25 years (SD = 3.5), with ages ranging from 7 to 18 years. All participants were male and not totally blind and had varying degrees of visual impairment, including partial vision, light perception, and various types of visual distortions. Six out of the eight participants had prior experience using AI applications such as ChatGPT or Copilot, whereas only two participants had previous experience using a white cane.

### 5.3.4 Questionnaire and Acceptance Measures
Participants were asked a set of questions both before and after the experiment, and oral consent for participation was also obtained. Before the experiment, participants were asked the following questions about their prior experience with any smart or non-smart white cane and any conversational AI applications:

- Have you ever used a white cane (a mobility cane for people who are visually impaired)?

- Have you ever used applications such as ChatGPT or other voice-based smart assistants?

After completing the tasks, participants answered a questionnaire based on the Unified Theory of Acceptance and Use of Technology (UTAUT) [39], focusing on their interaction with PISHYAR as a socially intelligent cane to measure their acceptance of using this new technology. The questionnaire includes these items: Anxiety (AN), Attitude Towards Technology (ATT), Facilitating Conditions (FC), Intention to Use (IU), Perceived Adaptiveness (PAD), Perceived Ease of Use (PEU), Perceived Usefulness (PU), Perceived Enjoyment (PE), Perceived Sociability (PS), Social Influence (SI), Social Presence (SP), and Trust (TR). The questionnaire statements are shown in Table 1. Responses were recorded using a 5-point Likert scale, and researchers documented the participants' answers.

**Table 1** Twelve common questionnaire statements adopted from UTAUT, rated orally by participants on a 5-point Likert scale after the experiment.

| No. | Item | Statement |
|---|---|---|
| 1 | AN | If I should use **PISHYAR**, I would be afraid to break something. |
| 2 | ATT | **PISHYAR** would make life more interesting. |
| 3 | FC | I know enough of **PISHYAR** to make good use of it. |
| 4 | IU | I'm certain to use **PISHYAR** in the future. |
| 5 | PAD | I think **PISHYAR** can be adaptive to what I need. |
| 6 | PEU | I find **PISHYAR** easy to use. |
| 7 | PU | I think **PISHYAR** can help me with many things. |
| 8 | PE | I enjoy doing things with **PISHYAR.** |
| 9 | PS | I find **PISHYAR** pleasant to interact with. |
| 10 | SI | I think it would give a good impression if I should use **PISHYAR.** |
| 11 | SP | I can imagine **PISHYAR** to be a living creature. |
| 12 | TR | I would follow the advice **PISHYAR** gives me. |

## 6 Results and Discussion

### 6.1 Evaluation of Socially Aware Navigation Performance

#### 6.1.1 Simulation-Based Evaluation of Obstacle Detection and Path Planning

Results of the simulation showed that in **nine out of ten scenarios**, the path planning algorithm successfully identified the optimal route and ensured correct guidance without accidents or obstacle contact. In only one case, due to specific environmental conditions and obstacle arrangement, the algorithm's performance required further optimization. Fig. 9 illustrates the different stages of the path planning process in one of these simulated scenarios, providing a clear representation of how the system detects obstacles and selects the appropriate route.

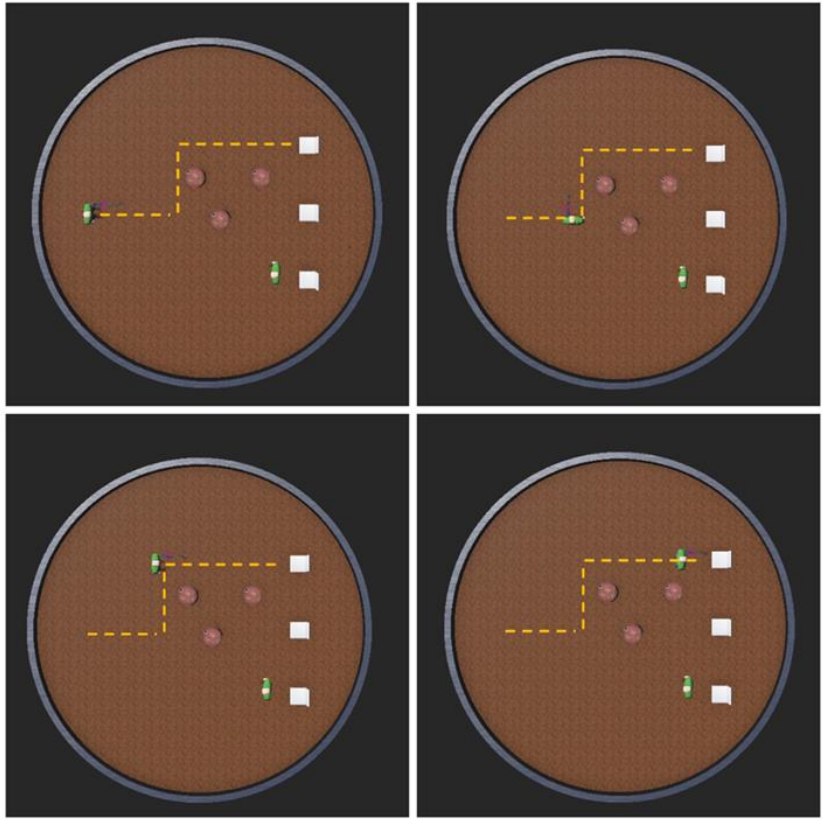

**Fig. 9** Representative stages of obstacle detection and path planning in the Webots simulation environment. The humanoid model equipped with the smart cane navigates toward the target chair while avoiding static obstacles.

### 6.1.2 Training and Validation Results of the COMPOSER Architecture

From the entire dataset, approximately 70% (equivalent to 1,778 clips) was used for model training, while 30% (765 clips) was allocated for performance evaluation. The model achieved a Top-1 accuracy of **95.7%**, which is consistent with the reported result in the original paper [36]. The confusion matrix of the model is shown in Fig. 10.

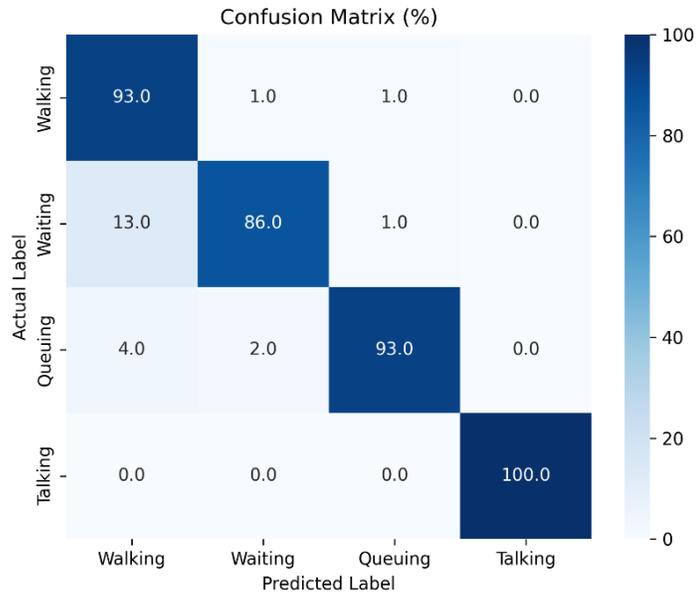

**Fig. 10** Confusion matrix of the COMPOSER architecture evaluated on the Collective Activity dataset, showing classification performance across walking, waiting, queuing, and talking activities.

Analysis of this matrix shows that the model achieved a flawless performance in classifying the talking class, reaching an accuracy of 100%. The highest level of confusion occurred between the walking and waiting classes, where in a few cases, samples from these two classes were misclassified. The classification accuracy for the queueing class was also very high, at 93%.

These results demonstrate the remarkable efficiency of the COMPOSER model in recognizing collective activities in diverse and complex environments. The only major challenge lies in distinguishing between behaviorally similar activities such as walking and waiting, which could be improved by enhancing the training dataset or employing more advanced preprocessing techniques. Overall, the developed model shows strong potential for practical applications and real-world scenarios.

### 6.1.3 Real-World Evaluation of the COMPOSER Architecture

Fig. 11 presents the COMPOSER model's accuracy in each of the 20 experimental conditions, providing a clear visual representation of its strengths and weaknesses in different scenarios and enabling a more detailed analysis of successes and prediction errors in each test.

The results of this evaluation showed that in **17 out of 20 scenarios**, the model was able to correctly predict the group activity present in the scene indicating acceptable accuracy even under challenging conditions (In all 20 scenarios, the correct label appeared among the model's top three predictions). Among the three misclassifications, one case involved an incorrect recognition of the waiting class, while the other two involved incorrect recognition of the walking class. In all these cases, the two classes were substituted for each other—a predictable and justifiable outcome given the behavioral and motion similarities between these activities. Similar misclassification patterns were also observed in the model's training results on the Collective Activity dataset.

Considering the limited number of experiments and the complexity of real-world conditions, the overall accuracy of **85%** obtained in this evaluation can be regarded as satisfactory for practical applications, highlighting the model's strong potential for integration into intelligent assistive systems.

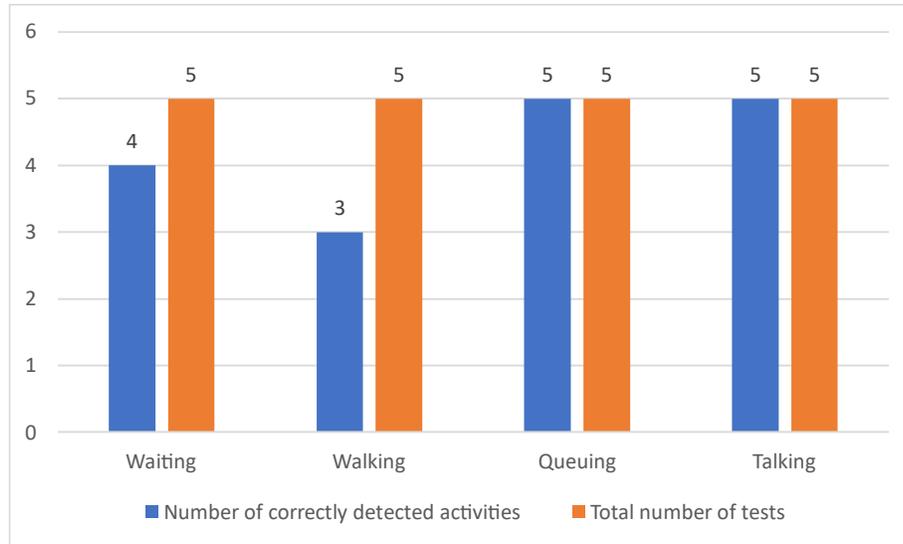

**Fig. 11** Accuracy of the COMPOSER architecture across 20 real-world experimental scenarios, illustrating correct and incorrect group activity predictions under varying environmental conditions.

### 6.1.4 System-Level Performance Evaluation of PISHYAR

The results obtained from evaluating PISHYAR in the three defined scenarios are as follows:

- **Scenario 1, Presence of Static Obstacles without Human Groups:** In this scenario, the system successfully completed all three trials. PISHYAR guided the user to the destination (vacant chair) without any collisions with static obstacles. These results demonstrate the stable performance and high accuracy of the system in environments with non-moving obstacles. The average time per trial in this scenario was approximately 25 seconds, indicating a fast system response.

- **Scenario 2, Static Obstacles with Changing Target Position Midway:** This scenario was more complex than the first. The system successfully completed 2 out of 3 trials, correctly finding an alternative path and guiding the user to the newly available vacant chair. In the unsuccessful trial, the primary cause was the user moving too close to the initial target chair, causing other chairs to move out of the depth camera's field of view. Additionally, persistent errors in data from the camera and IMU sensor prevented PISHYAR from correctly identifying alternative chairs and calculating the proper path. The average time per trial in this scenario was approximately 30 seconds, reflecting increased processing and decision-making time due to environmental complexity.

- **Scenario 3 – Social Navigation in the Presence of Human Groups:** This scenario was the most challenging, involving obstacles and human groups exhibiting different behaviors. For the *Talking* activity class, PISHYAR successfully guided the user in all three trials, maintaining proper social distance and etiquette. For the *Waiting* activity class (including *Queuing*), the system successfully navigated in 2 out of 3 trials; while in one trial, it mistakenly classified the activity as walking. In this case, since the system awaited the group to leave, the misclassification led to a temporary halt in navigation. For the *Walking* activity class, the system performed satisfactorily in 2 out of 3 trials, but in one trial, misclassification of the collective activity caused a collision with people in the path. Overall, out of 9 trials in this scenario, 7 were fully successful, corresponding to the estimated accuracy of **78%**. This performance is considered acceptable given the complex and dynamic social conditions. The average time per trial was approximately 35 seconds, slightly longer due to the complexity and multi-layered processing required. Fig. 12 summarizes the overall performance of the proposed system across all evaluation scenarios. This figure reflects the accuracy of reaching the target, avoiding obstacles, and maintaining social etiquette, clearly highlighting the strengths and limitations of the system under practical and real-world conditions.

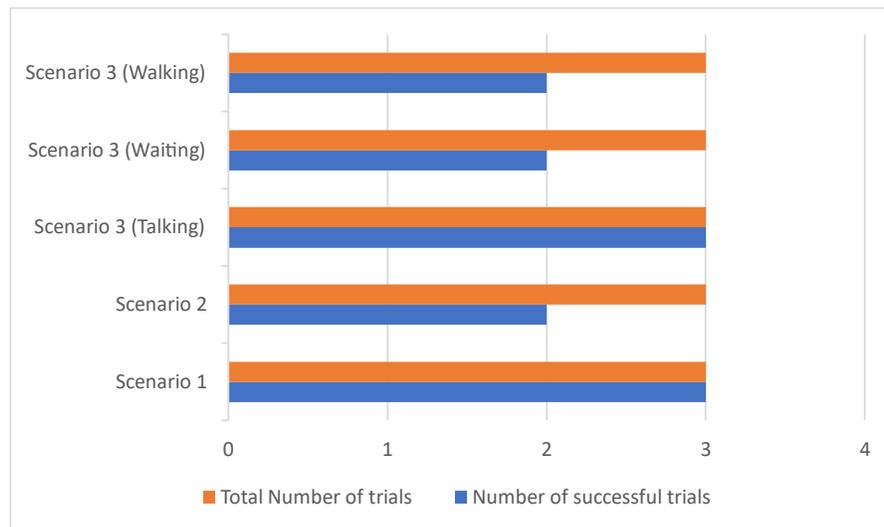

**Fig. 12** Summary of PISHYAR performance across all real-world evaluation scenarios, including obstacle avoidance, target reaching, and compliance with social navigation constraints.

The above results indicate that the proposed social navigation system on PISHYAR, despite certain limitations, has demonstrated acceptable performance in real-world environments with complex human conditions. The overall accuracy of the system across all evaluations was ~**80%**. Observed errors were mainly due to limited camera field of view, IMU sensor inaccuracies, and the lack of a comprehensive environmental map, which can be minimized through hardware improvements, enhanced positioning algorithms, and modifications to the navigation method. Additionally, the system's capability to detect and respect social etiquette, such as maintaining appropriate distance from human groups, highlights the innovation and practical applicability of this research, offering a notable advantage over many existing navigation systems.

To provide a benchmark, the results of the final evaluation in this study were compared with the data reported by Agrawal et al. [22]. Since social navigation was not addressed in their study or in much of the related literature, a direct comparison for social navigation is not possible. However, the primary objective of their work (finding an empty chair and navigating toward it in the presence of static obstacles) is closely aligned with the goals of this research, making it suitable for comparison. In addition to proximity, their criteria for selecting an empty chair also considered social aspects such as privacy, comfort, and social distance. The key difference is that in Agrawal et al.'s study, the social element was only incorporated in the chair selection process, not in the navigation itself. Their study conducted evaluations across 12 scenarios, with obstacles randomly placed in different positions and the goal being to reach a socially and spatially optimal chair without collisions. Their results showed that the system successfully achieved its objective in 10 out of 12 scenarios, corresponding to an overall accuracy of 83.3%.

This accuracy is comparable to the ~80% achieved in the present study, and the slight difference can be attributed to the added complexity of social navigation in this research. Social navigation introduces additional challenges compared to standard navigation with only static obstacles. It is also noteworthy that in scenarios involving only static obstacles, PISHYAR developed in this study successfully reached the target in all three trials, demonstrating full reliability under simpler conditions.

### 6.2 Evaluation of LLM-Based Assistive Interaction

#### 6.2.1 User Study Questionnaire Results

The results of the UTAUT questionnaire in our preliminary exploratory study for eight visually impaired participants are shown in Fig. 13. All questionnaire items were scored on a five-point Likert scale, with 1 indicating the minimum and 5 indicating the maximum score. In general, most items received mean scores of 4 or higher, which indicates a generally positive perception of PISHYAR among the participants. These results suggest a strong level of acceptance of the proposed system during our exploratory study which shows us a positive signal of our developed platform. Similar levels of user acceptance have been reported in recent assistive systems leveraging large language or vision language models for visually impaired users, including wearable VLM-based assistants [28] and real-time navigation aids [31].

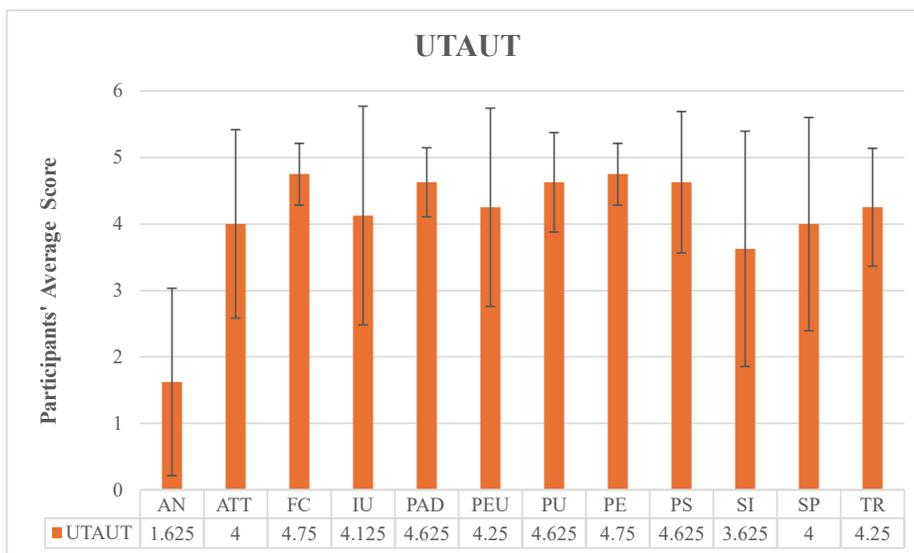

**Fig. 13** Mean participant scores for UTAUT questionnaire items (5-point Likert scale) evaluating acceptance, usability, trust, and perceived sociability of PISHYAR. Error bars indicate standard deviation.

Based on the results of our conducted study, the Anxiety (AN) item received a mean score of 1.625, which indicates a relatively low level of anxiety while using PISHYAR. This suggests that during interaction, participants generally did not feel afraid of breaking the system. This low anxiety level may indicate that the system was perceived as safe and user-friendly. However, the relatively high standard deviation for this item suggests individual differences, which may be related to participants' different levels of familiarity with technology.

As shown in Fig. 13, several items with high mean scores also exhibited relatively large standard deviations. For example, the Intention to Use (IU) item shows that while many participants expressed willingness to use PISHYAR again in the future, a smaller number were less inclined to do so. A similar pattern can be observed for Perceived Ease of Use (PEU), which suggests that although most participants found the system easy to use, some experienced difficulties during interaction. These difficulties may be related to system limitations, such as occasional uncertainty in responding to user inputs, which is common in LLM-powered interactive systems. The Pearson correlation coefficient between Anxiety (AN) and Perceived Ease of Use (PEU) was calculated as -0.97, highlighting a strong negative relationship. This result suggests that participants who found PISHYAR easier to use also reported lower levels of anxiety during interaction. This correlation is based on a tiny sample size and should not be considered as a conclusive statistical finding, but rather as an exploratory sign of connection.

The results for Facilitating Conditions (FC) indicate that participants felt they had sufficient knowledge to use PISHYAR. This may be explained by their prior experience with conversational AI systems such as ChatGPT or similar voice-based assistants. Based on the Perceived Adaptiveness (PAD) results, participants perceived PISHYAR as adaptive to their needs. In the context of this study, this suggests that the developed system and its integrated modules generally functioned as intended, with only minor malfunctions observed. The results for Perceived Enjoyment (PE) indicate that participants experienced enjoyment while using PISHYAR. In addition, the Perceived Sociability (PS) item suggests that the interaction was perceived as pleasant. Together, these findings highlight PISHYAR's role as a socially intelligent assistive cane that can provide a positive user experience.

The Trust item received a high score, which shows that participants generally perceived PISHYAR as trustworthy. Trust is a critical factor in human-robot interaction, particularly for assistive technologies. In contrast, Social Influence (SI), Social Presence (SP), and Attitude Toward Technology (AT) received comparatively lower scores than other questionnaire items. This suggests that, in its current form, PISHYAR may have a limited impact on users' social experiences or daily social life. However, the Perceived Usefulness (PU) item still received a high score, indicating that participants found the system helpful and functional.

The Pearson correlation coefficient between Intention to Use (IU) and Social Presence (SP) was calculated as 0.92, indicating a strong positive relationship. Since there were only a few participants, this correlation should be read cautiously and does not suggest that desire to use and perceived social presence are causally related. This result suggests that participants who were better able to perceive PISHYAR as a social entity were more likely to express willingness to use it again in the future. This finding highlights the importance of sociability and anthropomorphic cues in increasing long-term acceptance of socially assistive technologies. The strong correlation between perceived sociability and intention to use shows prior findings in socially assistive robotics, where anthropomorphic cues and conversational responsiveness were shown to directly influence long-term acceptance and trust [11, 25]. Unlike systems focusing solely on task efficiency, PISHYAR integrates conversational interaction into a physical mobility aid, which may increase the role of social presence in shaping user attitudes.

Overall, the questionnaire results indicate that PISHYAR is acceptable as a socially intelligent cane among visually impaired users; however, the observed trends are exploratory and should be validated in future studies with larger and more diverse populations. While its social impact is currently limited, the system shows potential to influence users' experiences positively with further improvements in social interaction capabilities.

**6.2.2 Interaction-Level System Performance**

Across all trials, PISHYAR successfully handled date and time requests as well as general question-and-answer commands issued by participants. The system also completed all scene description tasks without failure.

For object detection tasks involving "chair" and "bottle" detection, each of the eight participants performed both tasks, resulting in 16 detection trials in total. The combined VLM-YOLO pipeline successfully detected the target object in 15 out of 16 trials, with a single failure occurring in the water bottle detection. However, distance estimation in terms of step count for "chairs" was successfully provided in only 4 cases. Failures in distance estimation may be attributed to limitations in visual input, such as an incomplete view of the object or incorrect orientation of the cane relative to the target.

Interaction logs further indicate that during vision-based tasks, the system was also able to detect additional objects on the table, such as apples and cups. These objects were detected either by YOLO or the VLM, with the VLM being responsible for most successful detections. This suggests that the vision language model played a significant role in providing robust scene understanding beyond predefined object classes. This observation supports findings from recent multimodal foundation-model approaches, where VLMs demonstrated superior generalization beyond predefined object categories and provided richer scene understanding compared to standalone object detectors [27, 32]. Such flexibility is particularly valuable in assistive contexts, where the range of relevant objects cannot be predefined.

At the end of each session, participants were encouraged to ask PISHYAR additional questions beyond the structured tasks. Example queries included requests for information about the environment, such as "What is on the table in front of me?" and spatial guidance like "Can you tell me what is on my left side?" They also asked general knowledge and problem-solving questions, for instance, "What is ten times ten?" and personal advice, such as "My mother has a neck and back disc problem, what should I do to help her feel better?", "What should I do to improve my physical strength?", and "What should I do to improve my studies?" These interactions provided insights into the system's flexibility and participants' natural use patterns, which demonstrates its ability to handle unstructured, user-initiated requests in addition to task-oriented commands.

In contrast to robot-assisted platforms that rely on external computation or large mobile bases [30], PISHYAR demonstrates that multimodal LLM-based interaction can be effectively embedded within a lightweight cane-based form factor, albeit with trade-offs in sensing coverage and depth robustness.

Overall, the interaction-level results indicate that integrating LLMs and VLMs into a smart cane enables reliable conversational assistance and flexible scene understanding, while remaining constrained by camera viewpoint and sensing limitations. PISHYAR highlights both its useful potential and areas for further development by achieving a balance between portability and interaction richness as compared to prior wearable or robot-based systems.

## 7 Limitations and Future Work

This study faces several limitations related to sensing, navigation, interaction, and evaluation. In the social navigation component, one of the main limitations is the lack of an environmental map. The system relies solely on the RGB-D camera and IMU, which may reduce robustness in crowded or highly dynamic environments. This design choice was made to preserve real-time performance and to avoid requiring complex user movements during navigation. In addition, financial constraints prevented the use of high-cost sensors such as LiDAR. Furthermore, the COMPOSER-based group activity recognition model is limited to detecting a maximum of two human groups with up to 13 individuals in total. As a result, in dense crowds, the system may not fully capture all social interactions or enforce social navigation rules effectively.

Another limitation is related to the interaction module. Audio-based interaction, which depends on speech recognition and KWS, is sensitive to environmental noise, which could potentially affect recognition accuracy. In addition, the interaction mode routing mechanism, while effective in reducing computational load and power consumption, may introduce minor delays due to mode switching. This represents a trade-off between responsiveness and efficiency that is common in embedded assistive systems.

From an object perception perspective, the object detection and distance estimation module occasionally fail to detect certain obstacles. Since the YOLOv8 model trained on the COCO dataset was used, object recognition is limited to the 80 predefined classes in this dataset. Consequently, uncommon obstacles may not be detected reliably. Moreover, distance estimation accuracy depends on depth sensor data and may degrade under challenging lighting.

Finally, evaluation-related limitations should be noted. The relatively small size of the visually impaired population and the fact that all participants in the study were male may limit the generalizability of the findings across broader user groups with different gender, age groups, and cultural contexts. In addition, collective activity recognition is performed only at the beginning of navigation, assuming that no new human groups enter the environment during movement, which may limit system responsiveness in highly dynamic social settings.

Future work will focus on addressing these limitations through both hardware and algorithmic improvements. Planned extensions include the integration of advanced sensors such as LiDAR or higher-resolution depth cameras to improve perception accuracy, as well as the adoption of SLAM-based mapping and navigation to enable more reliable operation in dynamic and complex environments. The system will be extended to larger indoor spaces settings to increase its practical applicability. In addition, future versions will enhance multimodal feedback by combining both haptic and auditory cues, and improve user interaction through more user-friendly interfaces and organized instruction for users. On the interaction side, more robust STT and KWS models will be explored to increase reliability in noisy environments. From an evaluation perspective, future studies will involve a larger and more diverse participant group to improve generalizability. Furthermore, object detection performance can be improved by incorporating alternative detection models or segmentation-based approaches. Finally, future work will explore deeper integration of LLM- and VLM-based navigation using additional contextual inputs such as map information, as well as OCR or VLM-based text reading to support richer environmental understanding.

**8 Conclusion**

In this study, we designed and developed PISHYAR, a socially intelligent smart cane that combines socially aware navigation with multimodal LLM-based interaction to support both mobility and interactive assistance for visually impaired individuals in indoor environments. The system integrates three core components: (1) a hardware platform, including an OAK-D Lite RGB-D camera, a Raspberry Pi 5, and a modified white cane with a 3D-printed structure, three vibration motors for haptic feedback, and a power supply module; (2) a socially aware navigation framework that utilizes YOLOv8 for object detection, the D* Lite algorithm for path planning, and the COMPOSER transformer model for collective human activity recognition; and (3) an agentic LLM-VLM interaction framework that allows voice-based interaction, scene description, and object localization through GPT-4o, plus keyword spotting, speech-to-text, and text-to-speech modules.

The socially aware navigation component was evaluated through simulation and real-world experiments to measure its ability to guide users safely and respecting social etiquette at the same time. Across all navigation scenarios, the system achieved an overall success rate of approximately 80%. The system successfully combines obstacle avoidance, dynamic re-planning, and social compliance. In environments with only static obstacles, the cane achieved a 100% success rate, demonstrating stable baseline navigation. When target seats became occupied during movement, the system correctly re-planned in most cases, with failures primarily caused by the limited camera field of view and IMU drift. In socially dynamic scenarios involving walking, talking, and waiting groups, this system achieved seven out of nine successful trials by maintaining the proper interpersonal distance and modifying its behavior accordingly. These findings suggest that integration of the COMPOSER model enabled reliable recognition of collective human activities, with 85% accuracy in real-world tests. Overall, the results confirm that incorporating group activity awareness significantly enhances navigation quality compared to purely obstacle-based approaches.

The agentic LLM-based interaction framework was evaluated through structured user studies with visually impaired participants for measuring usability, acceptance, intention to use, trust, perceived usefulness, and socially relevant factors such as sociability and social presence. The system demonstrated stable and reliable performance in voice-based interaction, with all time/date queries and general conversational requests completed successfully. Scene description tasks were executed correctly in every trial, which indicates strong multimodal understanding. For object detection commands, the system correctly identified target objects in 15 out of 16 cases. However, distance estimation was only accurate in half of the trials, with errors mainly due to incomplete visual framing or cane misorientation. We have also performed a preliminary study for eight visually impaired participants using our developed social robot. Questionnaire results based on the UTAUT model revealed high user acceptance and strong perceptions of usefulness, trust, and sociability. Participants reported that PISHYAR was enjoyable to use during our experiments. Additionally, intention to reuse the system correlated strongly with perceived social presence. These findings suggest that the agentic interaction framework effectively supports natural and engaging assistance, while highlighting the need for improved spatial grounding and distance estimation in future iterations.

Overall, these results demonstrate that PISHYAR can effectively combine socially aware navigation and multimodal human-robot interaction to provide a user-friendly smart cane. The system provides visually impaired users with independent mobility in complex and socially dynamic contexts and provides the foundation for future works to integrate multimodal perception, navigation, and conversational AI into portable assistive devices.

**Acknowledgment**

This work is based upon research funded by the Sharif University of Technology. We would like to express our gratitude to our friends at the Social and Cognitive Robotics Laboratory. for cooperation and helps with the data collection, as well as all the participants. We would like to offer special thanks to Dr. Dezhan, Dr. Shirmohammadi, Mr. Asadi and the staff of Shahid Moheebi high-school for their cooperation during our experimental study.

**Statements & Declarations**

**Conflict of interest**

Author Alireza Taheri has received a research grant from the Sharif University of Technology. The authors Mahdi Haghighat Joo and Maryam Karimi Jafari declare that they have no conflict of interest.

**Availability of data and material (data transparency)**

All data from this project (the cane design files, codes, photos, videos, etc.) are available in the Social and Cognitive Robotics Laboratory archive.

**Code availability**

All of the codes are available in the archive of the Social and Cognitive Robotics Laboratory. In case the readers need the codes, they may contact the corresponding author.

**Authors' contributions**

**Consent to participate**

Informed consent was obtained from all individual participants included in the study.

**Appendix**

This appendix summarizes the prompts used in the proposed multimodal LLM-based interaction framework. The table presents the prompts employed by the LLM, the VLM, and the interaction routing module. The second column lists the prompt content, where ***System Instruction*** refers to the system-level prompt defining the model's role and behavior, and ***User Input*** corresponds to the user-role messages provided to the model during runtime, as defined in the OpenAI API message structure. The third column includes representative examples selected from experimental tests to illustrate how the prompts are instantiated in practice.

| Task | Prompt Template | Few-Shot Examples |
|---|---|---|
| **Interaction Mode Router (IMR)** | **system instruction:** You are a mode selector for a cane assistant. Decide if the user intent REQUIRES the camera/vision tools now.<br>Return STRICT JSON with:<br>{"mode": "VISION" \| "VOICE"}<br>Rules:<br>- Use "VISION" when the user asks about surroundings, people, objects, distances, obstacles, navigation, or anything needing the live camera.<br>- Use "VOICE" for date/time, chit-chat, general Q&A without current visual info.<br>- Be conservative: if camera is not clearly needed, pick "VOICE".<br>**user input:** (the output of the STT is sent as the input) | Input:<br>بطری آب را پیدا کن.<br>Output:<br>{"mode": "VISION"}<br><br>Input:<br>امروز چند شنبه است؟<br>Output:<br>{"mode": "VOICE"} |
| **Vision Task Router (VTR)** | **system instruction:** You are a routing controller for a multimodal cane assistant. Decide which TOOL to call based ONLY on the user's latest utterance.<br>TOOLS:<br>- SCENE: Use ONLY the VLM scene description (no object detector). Choose this when the user asks about surroundings /where they are/what's around/people/crowds/caution areas, or generally needs situational awareness.<br>- OBJECT: Use BOTH the VLM scene description AND the object detector. Choose this when the user asks to find/locate/identify a specific object or asks about nearby objects in general. If the user mentions a specific object (e.g., "cup", "empty chair", "phone"), extract it into `target`. If not, set `target` to null.<br>Return a STRICT JSON object with exactly these keys: | Input:<br>می‌تونی بگی که جلوی من چیه؟<br>Output:<br>{<br>  "action": "SCENE",<br>  "target": None<br>}<br><br>Input:<br>صندلی رو برام پیدا کن.<br>Output:<br>{<br>  "action": "OBJECT", |
| | `{`<br>  `"action": "SCENE" | "OBJECT",`<br>  `"target": "<string or null>"`<br>`}`<br>No extra text. No explanations. JSON only.<br><br>**user input:** (the output of the STT is sent as the input) |   `"target": "chair"`<br>`}` |
| **LLM** | **system instruction:** You are an interactive conversational assistant named 'Pishyār' (Persian: پیشیار), a friendly companion for the user. Introduce yourself briefly when appropriate, and speak with a warm and respectful tone. You may occasionally add a tiny touch of light humor or playfulness, but always stay concise and supportive. Avoid unnecessary complexity, extra commentary, or directly mentioning blindness unless the user explicitly does. Focus only on practical help and clear guidance in your responses, like a supportive friend.<br><br>**user input:**<br>You are a friendly and helpful companion for the user. Environmental information and text detected on signs:<br>{the VLM output}<br>The user has asked:<br>{the STT output}<br>Please provide a short, warm, fluent, and direct response in Persian. Do not start with greetings such as "hello" unless the user has greeted first. If text on a sign or board is detected, state it clearly and include it in the response if it helps with orientation or navigation. If distance information or the number of steps to nearby objects is available, be sure to include it so the user knows how far away objects are. Avoid unnecessary details, extra explanations, or any reference to the user being visually impaired. | **Input:**<br>You are a friendly and helpful companion for the user. Environmental information and text detected on signs:<br>{<br>توصیف صحنه: اتاقی با چیدمان ساده مشاهده می‌شود. در سمت چپ تصویر، بخشی از یک میز چوبی با یک بطری آب روی آن قرار دارد. دو صندلی با پشتی چوبی در نزدیکی میز دیده می‌شوند که به دیوار متصل شده‌اند. پرده‌های نازک بر روی پنجره‌ای که قسمت بالای تصویر را اشغال کرده، آویخته شده‌اند و مقداری نور از آن عبور می‌کند. در کنار صندلی‌ها، شوفاژ دیواری قرار دارد.<br>متن قابل خواندن روی تابلوها دیده نشد.<br>اشیای اطراف:<br>table: حدود ۲ قدم | chair: حدود ۴ قدم<br>جست‌وجو برای «chair»: پیدا شد<br>}<br>The user has asked:<br>{صندلی رو برام پیدا کن.}<br>Please provide a short, warm, fluent, and direct response in Persian. Do not start with greetings such as "hello" unless the user has greeted first. If text on a sign or board is detected, state it clearly and include it in the response if it helps with orientation or navigation. If distance information or the number of steps to nearby objects is available, be sure to include it so the user knows how far away objects are. Avoid unnecessary details, extra explanations, or any reference to the user being visually impaired.<br><br>**Output:**<br>صندلی‌ها حدوداً چهار قدم با تو فاصله دارند و در نزدیکی میز چوبی قرار گرفته‌اند. امیدوارم این اطلاعات کمکت کند! |
| **VLM** | **Text Prompt:** Look at the image and do TWO things:<br>1) Scene (Persian): Give a clear, natural Persian description (60–120 words) focusing on environment, key objects/people/actions, relative positions, movements, and useful distances. Do NOT mention colors.<br>2) OCR (fa/en): Extract all visible Persian or English sign/label/place-name text.<br>Return your answer in EXACTLY this template:<br>  Scene:<br>    <scene description in Persian><br>  Texts:<br>    - <lang:fa\|en> «{exact text}» | **Visual Input:**<br>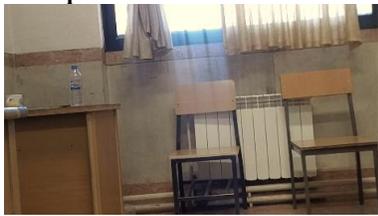<br><br>**Output:**<br>Scene: |

| | If there is no text, write:<br>   Texts:<br>     - none<br>**Visual Input:**<br>(An RGB image captured by the cane-mounted camera and provided to the model) | در این تصویر، اتاقی ساده دیده می‌شود. در سمت چپ، میز چوبی کوچک با سطحی صاف وجود دارد که روی آن یک بطری آب پلاستیکی و یک فنجان قرار گرفته‌اند. در میانه‌ی تصویر، دو صندلی چوبی با پایه‌های فلزی کنار هم در برابر دیواری با رادیاتور قرار گرفته‌اند. پرده‌هایی در بالا نصب شده‌اند که جلوی پنجره‌ای بزرگ را گرفته‌اند و نور از این پنجره‌ها به داخل اتاق می‌تابد.<br>Texts:<br>  - none |
|---|---|---|